\documentclass[11pt]{article}
\usepackage{graphicx} 
\usepackage{authblk}

\usepackage[letterpaper,top=2cm,bottom=2cm,left=3cm,right=3cm,marginparwidth=1.75cm]{geometry}

\usepackage{amsmath}
\usepackage{amssymb}
\usepackage{amsthm}
\usepackage{caption}

\usepackage{graphicx}    
\usepackage{subcaption}  
\usepackage{caption}     
\usepackage{float}  
\usepackage{hyperref}

\usepackage{comment}
\usepackage{dsfont}
\usepackage{makecell}

\usepackage[title]{appendix}

\usepackage{tikz}
\usetikzlibrary{arrows,automata,backgrounds,fit,patterns,petri,shadows,shapes,decorations.pathmorphing, bayesnet, calc}
\pgfdeclarepatternformonly{stripes}
{\pgfpointorigin}{\pgfpoint{0.25cm}{0.25cm}}
{\pgfpoint{0.25cm}{0.25cm}}
{
    \pgfpathmoveto{\pgfpoint{0cm}{0cm}}
    \pgfpathlineto{\pgfpoint{0.25cm}{0.25cm}}
    \pgfpathlineto{\pgfpoint{0.25cm}{0.12cm}}
    \pgfpathlineto{\pgfpoint{0.12cm}{0cm}}
    \pgfpathclose%
    \pgfusepath{fill}
    \pgfpathmoveto{\pgfpoint{0cm}{0.12cm}}
    \pgfpathlineto{\pgfpoint{0cm}{0.25cm}}
    \pgfpathlineto{\pgfpoint{0.12cm}{0.25cm}}
    \pgfpathclose%
    \pgfusepath{fill}
}

\tikzstyle{decision} = [rectangle, draw, thick, minimum size=1cm]
\tikzstyle{chance} = [circle, thick, minimum size = 1.15cm, inner sep= 1pt, draw = black, pattern = stripes, pattern color = gray!40]
\tikzstyle{chanceD} = [circle, thick, minimum size = 1.15cm, inner sep= 1pt, draw = black]
\tikzstyle{chanceD_dashed} = [circle, thick, minimum size = 1.15cm, inner sep= 1pt, draw = black, dashed]
\tikzstyle{chanceA} = [circle, thick, minimum size = 1.15cm, inner sep= 1pt, draw = black, fill = gray!40]
\tikzstyle{utility} = [regular polygon, regular polygon sides = 6, draw, thick, minimum size=1cm]
\tikzstyle{arrow} = [thick,->,>=stealth]
\tikzstyle{decision} = [rectangle, draw, thick, minimum size = 1.15cm, fill = gray!40]

\usepackage{amsthm}
\usepackage{thmtools}
\usepackage{amssymb} 
\declaretheorem[name=Example, qed=$\triangle$]{example}

\usepackage[square,numbers,sort]{natbib}

\title{A probabilistic framework for online test-time adaptation}

\author[a, b]{Daniel Corrales}
\author[a]{David Ríos Insua }

\affil[a]{Institute of Mathematical Sciences, ICMAT-CSIC, 28049 Madrid, Spain}
\affil[b]{Escuela de Doctorado, Universidad Autónoma de Madrid, 28049 Madrid, Spain}
\date{}

\begin{document}


\newcommand*\circled[1]{\tikz[baseline=(char.base)]{
            \node[shape=circle,draw,inner sep=2pt] (char) {#1};}}

\maketitle

\begin{abstract}
This paper presents a probabilistic framework for online test-time adaptation problems. In them, a model is trained on labeled data but must adapt to unlabeled data at test time under the assumption that training and test distributions potentially differ, that is, there might have been a distributional shift. The framework is based on a state-space modelling architecture from which parameter learning, parameter time evolution, prior tuning, and prediction can be characterized.
\end{abstract}

\section{Introduction and prior work}

One limitation of standard supervised discriminative learning \cite{bishop2006pattern} is that it assumes the environment is static after learning and, once the learning stage is over, the new incoming data is identically distributed, so the model will perform as well as it did in the learning phase. However, this is not always the case \cite{murphy2023probabilistic}. Some examples include recommender systems affected by changes in user behaviour, disease-monitoring models deployed in different hospitals on patients with different characteristics, or automatic driving systems as they circulate on roads under different atmospheric conditions \cite{koh2021wilds}, not to mention cases in which adversaries alter data with malicious purposes \cite{arce2025evasion}. To solve this issue, we may need to collect labeled data from the new environment and update and redo, at least, part of the learning process. Yet, both label collection and retraining are expensive tasks in terms of time and money. Thus, developing techniques to reduce such costs is desirable for the practical deployment of discriminative models in real-world scenarios. 

Although standard supervised discriminative learning is generally framed in such a way that it cannot leverage unlabelled data, several structural assumptions can be imposed in the learning process of a discriminative model to account for both labeled and unlabeled data simultaneously, resulting in the field of \emph{semi-supervised learning} \cite{seeger2000learning}. Some of these structural assumptions include \emph{smoothness} (if two data points in a high-density region are close, then so should be their corresponding labels); \emph{the manifold assumption} (high-dimensional data lie roughly on a low-dimensional manifold); or \emph{low-density separation} (for classification problems, decision boundaries should lie in low-density regions). Although these properties naturally arise in statistical learning, they actually become a design choice in semi-supervised learning.

Test-time adaptation \cite{sun2020test} deals with the situation in which we have a pretrained parametric model, $p( \mathbf{y}^S | \mathbf{x}^S,\Theta)$, parameterized by $\Theta$, trained using source labeled data $\mathcal{D}_S = \{ (\mathbf{x}_n^S, \mathbf{y}_n^S) \}_{n=1}^N$, and used for multi-class classification, together with an unlabeled dataset $\mathcal{X}^T = \{ \mathbf{x}^T_m\}_{m=1}^M$, with potential distributional shift from the data used in training. If this pretrained model could correctly predict over the whole data space, then a change in $p(\mathbf{x})$ would not affect the predictions in any way. However, data is usually scarce, and predictive models are accurate only in parts of the input space close to the training data. Therefore, they can be affected by covariate shift, that is, situations in which $p(\mathbf{x}^S) \neq p(\mathbf{x}^T)$. The main motivation of TTA is to adapt the parameters of a pretrained model to improve its predictive performance on the target domain, or on both source and target domains. Online TTA (OTTA) deals with the situation in which there is a temporal covariate shift, for which adaptation shall be done sequentially \cite{xiao2024beyond, liang2025comprehensive}.

The vast majority of works on TTA \cite{wang2021tent, zhang2022memo, niu2022efficient, marsden2024universal} employ \emph{entropy minimization} \cite{grandvalet2004semi} to leverage unlabeled data. This entails a low-density assumption that relies on the premise that all inputs $X$ should have a corresponding unique labelling $Y$, and thus the data aleatoric uncertainty should be low. Usually employed in the form of a loss function, entropy minimization separates clusters of features and forces the classification boundary to be as far from the clusters as possible. Using this unsupervised loss naively on any random parameter initialization will most likely not result in good predictive results. However, recall that, in TTA, this loss is used on a pretrained model. Therefore, the decision boundary is already a good fit, and, given similar but shifted data, this loss will find low-density separation regions to improve predictive capabilities in the new data. In modelling terms, most of these works deal with pretrained deep neural networks and, during adaptation, only a few of their parameters are adapted while the rest remain frozen. In particular, within the image domain, these unfrozen parameters usually include batch normalization and layer normalization parameters, but last-layer methods are also popular, as they greatly simplify inference.

Although much of the TTA literature focuses on deterministic methods, such as \cite{wang2021tent, marsden2024universal}, some works on TTA exist within a probabilistic setting. For example, \citet{zhou2021bayesian} characterize the data-generating process graphically and employ entropy minimization in the construction of a prior distribution. More recent works focus on a Bayesian filter for the noisy parameter updates resulting from using entropy minimization as a proxy loss \cite{lee2024continual, lee2024stationary, leebayesian}. Other relevant works involve using generative models to model cluster prototypes \cite{schirmer_temporal_2025}. However, there is still a significant literature gap on general probabilistic modelling of TTA problems, as most works bypass the full specification of the modelling decisions without many arguments.

This work develops a probabilistic OTTA methodology for sequential data that provides uncertainty estimates on predictions, in a similar fashion to previous work on probabilistic online learning \cite{duran2024unifying}. Our main contributions are the following. First, we propose a fully probabilistic setting for classification OTTA classification, in which uncertainty can be quantified at all steps of the pipeline. Second, we characterize each of the building blocks leveraging concepts from dynamic linear models \cite{west1997bayesian} and Gibbs posteriors \cite{bissiri2016general}. Finally, we approximate these posteriors using Gaussian distributions for real-time sequential Bayesian filtering and decision-support \cite{jones2024bayesian, duran2026martingale}. We argue that this probabilistic approach serves as an umbrella for a vast collection of state-of-the-art TTA methods as special cases and design choices of structural elements of this framework, while also proposing novel competitive TTA techniques.

\section{OTTA problem set up}

Let $\mathcal{D}_S = \{ (x_n^S, y_n^S) \}_{n=1}^N$ denote the source dataset used to pretrain a model with parameter $\boldsymbol{\theta}_S \in \Theta_S$, and let $\boldsymbol{\theta}_t \in \Theta_t$, with $t \in \mathbb{N}$, denote the parameters that are adapted sequentially based on test data at time $t$. Both parameter sets, $\Theta_S $ and $\Theta_t$, may coincide (e.g., the parameters of a logistic regression model), one may be a subset of the other (e.g., adapt the last layer of a pretrained neural network), or they might be independent from each other, each set representing different types of parameters (e.g., prototype based model that uses source neural features). We do assume for simplicity that the parameter set at test-time $\Theta_t$ is constant for all $t$.  At test time, we observe an unlabeled sequence $\{ \mathcal{X}_t \}_{t \in \mathbb{N}}$ drawn from target distributions that may differ from the source, while the corresponding label sequence $\{ \mathcal{Y}_t \}_{t \in \mathbb{N}}$ is never observed.

The objective in OTTA is to predict $\mathbf{y}_T \in \mathcal{Y}_T$ at any time T for each $\mathbf{x}_T \in \mathcal{X}_T$ using information accumulated up to time $\tau(T)$, where $\tau(T) \in \{1,...,T\}$ is a function specifying which test observations are used to adapt the parameters before predicting at time $t$. The predictive choice $\tau(T) = T-1$, in which the output at any time depends only on past inputs, is the standard in online learning but discards the distributional information that $\mathbf{x}_T$ carries about the current state of $\boldsymbol{\theta}_T$. Under covariate shift, $\mathbf{x}_T$ carries information about the current parameter state $\boldsymbol{\theta}_T$ but is conditionally independent of $\mathbf{y}_T$ given $\boldsymbol{\theta}_T$ in the measurement model, so introducing it into the filtering posterior before predicting may be justified as it adds information with no label leakage. This is the main argument behind filter choice $\tau(T) = T$ in TTA, where the output at any time depends on the current and past inputs. This is implicitly assumed in most of the methods in the literature and has connections with transductive learning \cite{chapelle_semi-supervised_2006}.

\begin{figure}[h!]
    \begin{center}
    \begin{tikzpicture}[node distance=2.2cm]

        \node (D_source) [decision]{$\mathcal{D}_S$};
        \node (theta_S) [chanceD, below of = D_source]{$\boldsymbol{\theta}_S$};
        \node (theta_0) [chanceD, below of = theta_S] {$\boldsymbol{\theta}_{0}$};

        \node (theta_1) [chanceD, right of = theta_0]{$\boldsymbol{\theta}_{1}$};
        \node (y_1) [chanceD, above of = theta_1] {$\mathbf{y}_{1}$};
        \node (x_1) [chanceA, above of = y_1] {$\mathbf{x}_{1}$};

        \node (dots1) [right of = x_1]{$\hdots$};
        \node (dots2) [right of = theta_1]{$\hdots$};
        \node (dots3) [right of = y_1]{$\hdots$};

        \node (x_t-1) [chanceA, right of = dots1] {$\mathbf{x}_{t-1}$};
        \node (y_t-1) [chanceD, below of = x_t-1] {$\mathbf{y}_{t-1}$};
        \node (theta_t-1) [chanceD, below of = y_t-1]{$\boldsymbol{\theta}_{t-1}$};
        
        \node (x_t) [chanceA, right of = x_t-1] {$\mathbf{x}_t$};
        \node (y_t) [chanceD, below of = x_t] {$\mathbf{y}_t$};
        \node (theta_t) [chanceD, below of = y_t] {$\boldsymbol{\theta}_t$};
        
        \node (dots1_1) [right of = x_t]{$\hdots$};
        \node (dots2_1) [right of = theta_t]{$\hdots$};
        \node (dots3_1) [right of = y_t]{$\hdots$};
    
        \draw[arrow](D_source) to (theta_S);
        \draw[arrow](theta_S) to (theta_0);
        \draw [arrow] (theta_0) to (theta_1);
        
        \draw [arrow, bend right = 35, dashed] (theta_1) to (x_1);
        \draw [arrow] (theta_1) to (y_1);
        \draw [<->, >=stealth, thick] (x_1) to (y_1);
        \draw [arrow] (theta_1) to (dots2);

        \draw [arrow] (dots2) to (theta_t-1);
        \draw [arrow, bend right = 35, dashed] (theta_t-1) to (x_t-1) ;
        \draw [arrow] (theta_t-1) to (y_t-1);
        \draw [<->, >=stealth, thick] (x_t-1) to (y_t-1);
        \draw [arrow] (theta_t-1) to (theta_t);

        \draw [arrow, bend right = 35, dashed] (theta_t) to (x_t);
        \draw [arrow] (theta_t) to (y_t);
        \draw [<->, >=stealth, thick] (x_t) to (y_t);
        \draw [arrow] (theta_t) to (dots2_1);
    
        \node[draw, rectangle, rounded corners, fit=(x_1)(y_1), inner sep=5pt] (groupB) {};
        \node[draw, rectangle, rounded corners, fit=(x_t-1)(y_t-1), inner sep=5pt] (groupN_1) {};
        \node[draw, rectangle, rounded corners, fit=(x_t)(y_t), inner sep=5pt] (groupN) {};
    
        \node at (groupB.west) [right] {$N_{1}$};
        \node at (groupN_1.west) [right] {$N_{t-1}$};
        \node at (groupN.west) [right] {$N_t$};
    
    \end{tikzpicture}
    \end{center}
    \caption{TTA probabilistic model. Grey nodes represent observed variables. White nodes, unobserved ones. Circled nodes are random variables, while squared nodes are deterministic for TTA. Rounded boxes imply that data may be coming in batches. Solid arcs represent true conditionals. Dashed arcs represent the Gibbs observation model.}
    \label{fig:OTTA_graphical_model}
\end{figure}
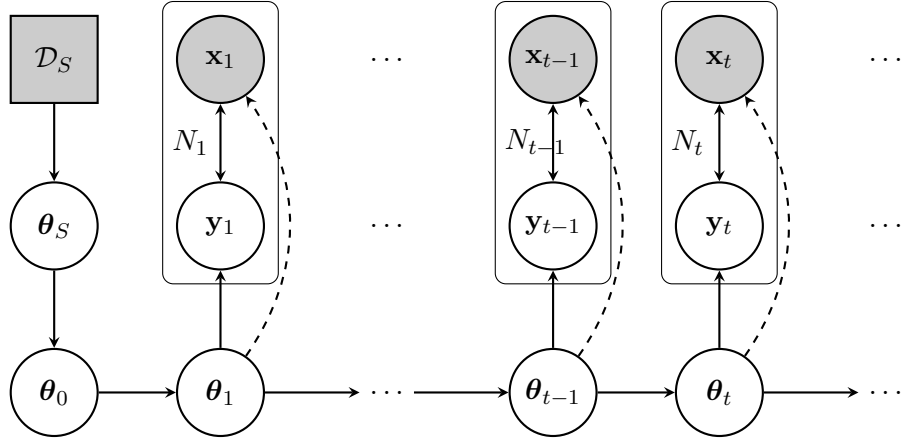

Figure \ref{fig:OTTA_graphical_model} represents the graphical model defining our probabilistic OTTA framework. Proceeding temporally, the first nodes define a conditional distribution $p(\boldsymbol{\theta}_S | \mathcal{D}_S)$, which is the pretrained model posterior on $\boldsymbol{\theta}_S$ after observing source labeled data $\mathcal{D}_S$. After that, adaptation parameters $\boldsymbol{\theta}_t$ are initialized at $t=0$ based on their architectural relationship with $\boldsymbol{\theta}_S$. At any time $t$, we have a classifier with joint form $p(\mathcal{X}_t, \mathcal{Y}_t | \boldsymbol{\theta}_t)$. As $\mathcal{Y}_t$ is not available, our observation model will be $p(\mathcal{X}_t | \boldsymbol{\theta}_t)$, which will be characterized either by a latent variable model or by a loss-based model. Importantly, even though $\mathcal{Y}_t$ is not available at time $t$, we will be interested in prediction and thus a predictive model $p(\mathcal{Y}_t | \mathcal{X}_t, \boldsymbol{\theta}_t)$ will also need to be specified. As a last step, as we are in a sequential setting, we will also need to characterize a transition for the parameters of the form $p(\boldsymbol{\theta}_t | \boldsymbol{\theta}_{t-1})$. Table \ref{tab:building_blocks} provides a summary of the necessary building blocks for this framework, which will be characterized in the following section. Observe that at time $T$, the joint parameter posterior will be factorized as
\begin{align}
    p(\boldsymbol{\theta}_S, \boldsymbol{\theta}_{0:T} | \mathcal{X}_{1:T}, \mathcal{D}_S) \propto p(\boldsymbol{\theta}_S | \mathcal{D}_S) \cdot p(\boldsymbol{\theta}_0 | \boldsymbol{\theta}_S) \cdot \prod_{t=1}^T p(\boldsymbol{\theta}_t | \boldsymbol{\theta}_{t-1}) \cdot p(\mathcal{X}_t | \boldsymbol{\theta}_t)
\end{align}

We may also define the propagation, updating, and predicting schemes at each time step. Proceeding temporally, we propagate the prior uncertainty at $t-1$, denoted by $p(\boldsymbol{\theta}_{t-1}| \mathcal{X}_{1:t-1}, \mathcal{D}_S)$, using the transition function $p(\boldsymbol{\theta}_t| \boldsymbol{\theta}_{t-1})$ to obtain a prior distribution at time $t$
\begin{align}\label{eq:prior}
    p(\boldsymbol{\theta}_t|  \mathcal{X}_{1:t-1}, \mathcal{D}_S) &=  \int p(\boldsymbol{\theta}_t| \boldsymbol{\theta}_{t-1}) \cdot p(\boldsymbol{\theta}_{t-1}|  \mathcal{X}_{1:t-1}, \mathcal{D}_S)  \ d\boldsymbol{\theta}_{t-1} 
\end{align}
At this time step, we only observe $ \mathcal{X}_t$ (and not $\mathcal{Y}_t)$. The updated posterior for $\boldsymbol{\theta}_t$ will then be
\begin{align}\label{eq:unsupervised_posterior}
    p(\boldsymbol{\theta}_t | \mathcal{X}_{1:t}, \mathcal{D}_S) &\propto p(\mathcal{X}_t | \boldsymbol{\theta}_t) \cdot p(\boldsymbol{\theta}_t |  \mathcal{X}_{1:t-1}, \mathcal{D}_S) 
\end{align}

We shall be interested in doing same-step prediction, as we have not observed any $\mathcal{Y}_t$; therefore, we predict labels through
\begin{align}
    p(\mathcal{Y}_t|  \mathcal{X}_{1:t}, \mathcal{D}_S)
        &= \int p(\mathcal{Y}_t| \mathcal{X}_{t},\boldsymbol{\theta}_{t}) \cdot p(\boldsymbol{\theta}_{t} | \mathcal{X}_{1:t}, \mathcal{D}_S) d\boldsymbol{\theta}_t
\end{align}
Note that doing prediction in this way is not strictly double-counting the data, as the observation model does not depend on $\mathcal{Y}_t$, which is unobserved. This approach has the taste of transductive learning \cite{chapelle_semi-supervised_2006}, as the posterior of $\boldsymbol{\theta}_t$ is conditioned on $X_t$, and then used to predict $Y_t$, implying that the goal might not necessarily be to find a general mapping between any $X$ and $Y$, but a specific mapping between $X_t$ and $Y_t$. However, as we repeat this procedure in an online fashion for different $X_t$, we assume that the posterior of $\boldsymbol{\theta}_t$ accumulates information about all the previous steps and therefore may generalize well over all $X$. As we will see, this is not guaranteed and can lead to instabilities or catastrophic forgetting issues, so we will need to find ways to control for this.

\begin{table}[h!]
    \centering
    \begin{tabular}{|c |c |}
    \hline
        \textbf{Building block} & \textbf{Description} \\ \hline
        $p(\boldsymbol{\theta}_S | \mathcal{D}_S)$ & Source posterior \\ \hline
        $p(\boldsymbol{\theta}_0 | \boldsymbol{\theta}_S)$ & Adaptation parameter initialization \\ \hline
        $p(\mathcal{Y}_t | \mathcal{X}_t, \boldsymbol{\theta}_t)$ & Predictive model    \\ \hline
        $p(\mathcal{X}_t | \boldsymbol{\theta}_t)$ & Observation model   \\ \hline
         $ p(\boldsymbol{\theta}_t | \boldsymbol{\theta}_{t-1})$ & Parameter transition  \\ \hline 
    \end{tabular}
    \caption{Building blocks for OTTA characterization}
    \label{tab:building_blocks}
\end{table}

\section{Model characterization}

As Table \ref{tab:building_blocks}, the model gets fully characterized by choosing how the design of the source posterior, the adaptation parameter initialization, the predictive model, the observation model, and the parameter transition.

\subsection{\texorpdfstring{Source posterior $p(\boldsymbol{\theta}_S | \mathcal{D}_S)$}{Source posterior}}\label{sec:source_posterior}
In the standard TTA literature, fundamentally based on point-wise optimization methods, the source posterior is approximated via a single optimal parameter $\widehat{\theta}$, which translates to using a point-mass distribution \begin{align}
p(\boldsymbol{\theta}_S | \mathcal{D}_S) = \delta(\boldsymbol{\theta}_S - \boldsymbol{\theta}_S^*).
\end{align}
This is straightforward as $\boldsymbol{\theta}_S^*$ corresponds to the optimal parameter after training a point-estimate model, and is computationally free beyond training, as it does not require any additional computation to estimate uncertainty. Many of the state-of-the-art works on image classification via neural networks (NNs)  \cite{wang2021tent, niu2022efficient, lee2024continual, marsden2024universal} make this assumption. 

However, a better approximation of the source posterior could be used, even for complex intractable distributions, as when using NNs. A principled tractable choice for approximating the source posterior, valid even for deterministic models, is the Laplace approximation \cite{mackay1992practical}. Here, the source posterior is approximated by a Gaussian 
\begin{align}\label{eq:laplace_approx_source_hessian}
    p(\boldsymbol{\theta}_S | \mathcal{D}_S) \approx \mathcal{N}(\theta^*_{S}, \Lambda_S^{-1})
\end{align} where $\theta^*_{S}$ is the maximum a posteriori (MAP) solution and $\Lambda_S^{-1}$ is the Hessian of the negative log posterior evaluated at the MAP, which may be approximated to ensure its positive definiteness \cite{daxberger2021laplace}. This source posterior approximation is relatively standard in the continual learning literature. It connects directly with Elastic Weight Consolidation \cite{kirkpatrick2017overcoming}, a technique to avoid catastrophic forgetting in continual problems with NNs, essentially regularizing learning through the source posterior.

If the pretrained model is a probabilistic model itself, then we would already have an approximation of the actual source posterior that we could use for adaptation, and we would not require post-hoc methods such as the Laplace approximation. This is the case, for example, of Bayesian logistic regression, Bayesian kernel methods, or Bayesian NNs.

\subsection{\texorpdfstring{OTTA parameter initialization $p(\boldsymbol{\theta}_0 | \boldsymbol{\theta}_S)$}{OTTA parameter initialization}}
Given a choice of source posterior, the initialization $p(\boldsymbol{\theta}_0 | \boldsymbol{\theta}_S)$ specifies the distribution over adaptable parameters before any test observation arrives. Its specification first requires establishing the structural relationship between $\boldsymbol{\theta}_S$ and $\boldsymbol{\theta}_t$, for which two main scenarios arise.

When $\Theta_t \subseteq \Theta_S$, the adaptable parameters are a subset of the source ones. The extreme case where $\Theta_t = \Theta_S$ adapts the entire model, thus maximising adaptation flexibility.  Restricting to a strict subset $\Theta_t \subset \Theta_S$ trades adaptation flexibility for computational tractability and is the approach followed by most TTA works, which may adapt only batch/layer normalization parameters \cite{wang2021tent} or last layer parameters \cite{iwasawa2021test, schirmer_temporal_2025} of a neural network. The initialization is then the marginal of the source posterior of Section~\ref{sec:source_posterior} over $\Theta_t$,
\begin{align}
    p(\boldsymbol{\theta}_0 \mid \boldsymbol{\theta}_S) = \int p(\boldsymbol{\theta}_S \mid \mathcal{D}_S) \, d\boldsymbol{\theta}_{S \setminus \boldsymbol{\theta}_0},
\end{align}
where $\boldsymbol{\theta}_{S\setminus 0}$ denotes the non-adaptable source parameters. For a deterministic source, this marginal is a point mass at $\boldsymbol{\theta}_{S\mid\boldsymbol{\theta}_0}^*$, the components of the source MAP belonging to the adaptable subspace. For a Laplace or otherwise Gaussian source, it is the Gaussian block $\mathcal{N}(\boldsymbol{\theta}_{S\mid\boldsymbol{\theta}_0},\, \Lambda_{S\mid\boldsymbol{\theta}_0}^{-1})$, with $\boldsymbol{\theta}_{S\mid\boldsymbol{\theta}_0}$ and $\Lambda_{S\mid\boldsymbol{\theta}_0}^{-1}$ the corresponding blocks of Equation (\ref{eq:laplace_approx_source_hessian}), restricted to the adaptable subspace.

When $\Theta_t \cap \Theta_S = \emptyset$, the adaptable parameters are introduced at test time with no source counterpart. This is the case for last-layer methods based on mixture models \cite{schirmer_temporal_2025}, where parameters model class-prototype dynamics, and for low-rank adaptation (LoRA) techniques \cite{hu2022lora, kojima2025lora}, where auxiliary parameters are added to frozen source parameters via low-rank coupling. With no source posterior to marginalise, $p(\boldsymbol{\theta}_0 \mid \boldsymbol{\theta}_S)$ must be specified independently --- typically as a point mass at a chosen initial value (e.g., zero for LoRA adapters) or a broad Gaussian prior.

\subsection{\texorpdfstring{Predictive model $p(y_t| x_t, \boldsymbol{\theta}_t)$}{Predictive model}}
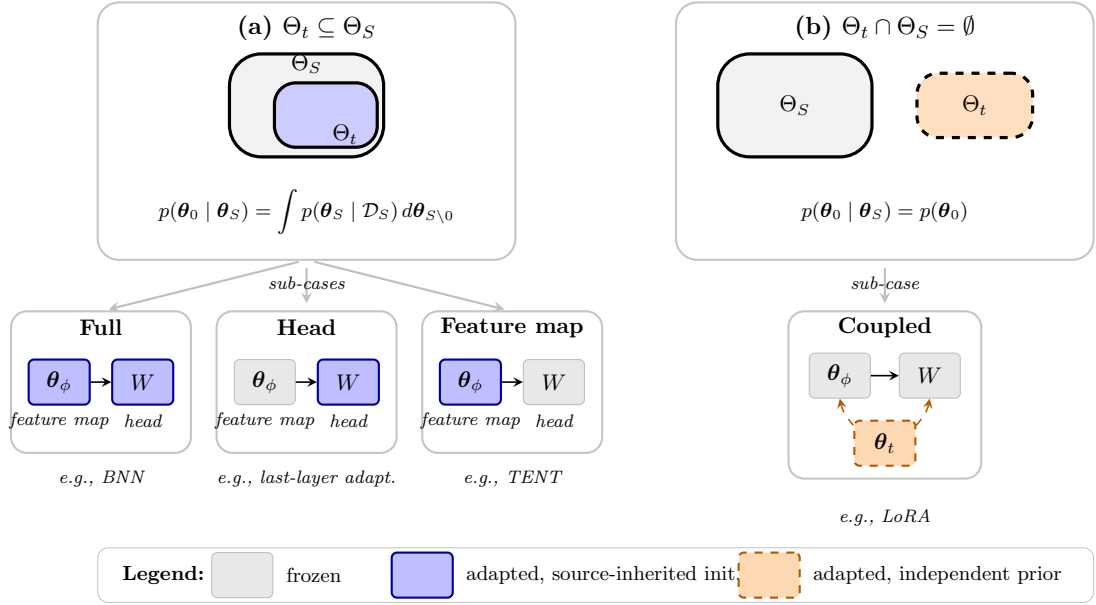
\begin{figure}[t]
\centering
\begin{tikzpicture}[scale=0.85, transform shape,
    setS/.style={draw, very thick, fill=gray!10, rounded corners=12pt,
                 minimum width=2.4cm, minimum height=1.6cm},
    setT/.style={draw, very thick, fill=blue!20, rounded corners=8pt},
    setTaux/.style={draw, very thick, fill=orange!25, dashed, rounded corners=8pt},
    topframe/.style={draw=gray!45, thick, rounded corners=8pt,
                     minimum width=6.5cm, minimum height=4.0cm, inner sep=6pt},
    subpanel/.style={draw=gray!45, thick, rounded corners=6pt,
                     minimum width=3.0cm, inner sep=6pt},
    slot/.style={draw, rectangle, rounded corners=2pt,
                 minimum width=0.95cm, minimum height=0.7cm,
                 font=\small, inner sep=2pt},
    frozen/.style={slot, fill=gray!18, draw=gray!50},
    adapted/.style={slot, fill=blue!25, thick, draw=blue!60!black},
    aux/.style={slot, fill=orange!30, thick, dashed, draw=orange!70!black},
    toplabel/.style={font=\normalsize\bfseries},
    sublabel/.style={font=\small\bfseries},
    eqannot/.style={font=\footnotesize},
    slotlabel/.style={font=\scriptsize\itshape},
    methodlabel/.style={font=\scriptsize\itshape},
    arr/.style={->, >=stealth, semithick},
    couple/.style={->, >=stealth, semithick, dashed, draw=orange!70!black},
    expand/.style={->, >=stealth, thick, gray!50, shorten >=3pt, shorten <=3pt},
    expandlabel/.style={font=\scriptsize\itshape},
]


\node[topframe] (FL) at (-4.5, 4.3) {};
\node[toplabel] at (-4.5, 5.9) {(a) $\Theta_t \subseteq \Theta_S$};

\node[setS] (SL) at (-4.5, 4.7) {};
\node[setT, minimum width=1.6cm, minimum height=1.0cm] at (-4.2, 4.55) {};
\node[font=\small] at (-4.5, 5.35) {$\Theta_S$};
\node[font=\small] at (-3.9, 4.25) {$\Theta_t$};

\node[eqannot] at (-4.5, 3.05) {%
    $p(\boldsymbol{\theta}_0 \mid \boldsymbol{\theta}_S) =
     \displaystyle\int p(\boldsymbol{\theta}_S \mid \mathcal{D}_S)\, d\boldsymbol{\theta}_{S\setminus 0}$%
};

\node[topframe] (FR) at (4.5, 4.3) {};
\node[toplabel] at (4.5, 5.9) {(b) $\Theta_t \cap \Theta_S = \emptyset$};

\node[setS] (SR) at (3.1, 4.7) {};
\node[setTaux, minimum width=1.8cm, minimum height=1.0cm] at (5.9, 4.7) {};
\node[font=\small] at (3.1, 4.7) {$\Theta_S$};
\node[font=\small] at (5.9, 4.7) {$\Theta_t$};

\node[eqannot] at (4.5, 3.05) {%
    $p(\boldsymbol{\theta}_0 \mid \boldsymbol{\theta}_S) = p(\boldsymbol{\theta}_0)$%
};


\draw[expand] (-4.5, 2.3) -- (-7.7, 1.5);
\draw[expand] (-4.5, 2.3) -- (-4.5, 1.5);
\draw[expand] (-4.5, 2.3) -- (-1.3, 1.5);
\draw[expand] (4.5, 2.3) -- (4.5, 1.5);

\node[expandlabel] at (-4.5, 1.95) {sub-cases};
\node[expandlabel] at (4.5, 1.95) {sub-case};


\foreach \id/\xc/\name/\phistyle/\Wstyle/\methodname in {%
    A/-7.7/Full/adapted/adapted/{e.g., BNN},
    B/-4.5/Head/frozen/adapted/{e.g., last-layer adapt.},
    C/-1.3/{Feature map}/adapted/frozen/{e.g., TENT}%
} {%
    \node[subpanel, minimum width=2.8cm, minimum height=2.2cm] at (\xc, 0.4) {};
    \node[sublabel] at (\xc, 1.25) {\name};
    \node[\phistyle] (phi\id) at (\xc-0.65, 0.4) {$\boldsymbol{\theta}_\phi$};
    \node[\Wstyle]   (W\id)   at (\xc+0.65, 0.4) {$W$};
    \draw[arr] (phi\id) -- (W\id);
    \node[slotlabel] at (\xc-0.65, -0.2) {feature map};
    \node[slotlabel] at (\xc+0.65, -0.2) {head};
    \node[methodlabel] at (\xc, -1.1) {\methodname};
}

\node[subpanel, minimum height=2.6cm] at (4.5, 0.2) {};
\node[sublabel] at (4.5, 1.25) {Coupled};
\node[frozen] (Cphi) at (3.8, 0.5) {$\boldsymbol{\theta}_\phi$};
\node[frozen] (CW)   at (5.2, 0.5) {$W$};
\draw[arr] (Cphi) -- (CW);
\node[aux] (Caux) at (4.5, -0.55) {$\boldsymbol{\theta}_t$};
\draw[couple] (Caux) to[bend left=15]  (Cphi.south);
\draw[couple] (Caux) to[bend right=15] (CW.south);
\node[methodlabel] at (4.5, -1.7) {e.g., LoRA};


\node[draw=gray!45, rounded corners=4pt, inner sep=6pt,
      minimum width=15.5cm, minimum height=0.85cm] at (0, -2.6) {};
\node[font=\footnotesize\bfseries, anchor=west] at (-7.5, -2.6) {Legend:};

\node[frozen] at (-5.5, -2.6) {};
\node[font=\footnotesize, anchor=west] at (-4.95, -2.6) {frozen};

\node[adapted] at (-2.7, -2.6) {};
\node[font=\footnotesize, anchor=west] at (-2.15, -2.6) {adapted, source-inherited init.};

\node[aux] at (2.7, -2.6) {};
\node[font=\footnotesize, anchor=west] at (3.25, -2.6) {adapted, independent prior};

\end{tikzpicture}
\caption{Structural relationships between source parameters $\boldsymbol{\theta}_S$ and adaptable parameters $\boldsymbol{\theta}_t$. \textbf{Top:} two cases distinguished by the set-theoretic relation between $\Theta_S$ and $\Theta_t$, with the corresponding initialisation $p(\boldsymbol{\theta}_0 \mid \boldsymbol{\theta}_S)$ shown beneath each. \textbf{Bottom:} sub-cases distinguished by which component of the predictive model carries $\boldsymbol{\theta}_t$. In (b), disjointness refers to parameter structure, that is, the auxiliary parameters are not in $\boldsymbol{\theta}_S$. However, they are not functional independent, as the dashed coupling arrows indicate.}
\label{fig:structural_cases}
\end{figure}

The predictive model $p(y_t| x_t, \boldsymbol{\theta}_t)$ specifies how predictions are formed from the current parameter state. For classification with $K$ classes, it adopts the general form
\begin{align}\label{eq:predictive_model}
    p(Y | \mathcal{X}_t, \boldsymbol{\theta}_t) = \text{Categorical}(\text{Softmax}(f(\mathcal{X}_t; \boldsymbol{\theta}_t))),
\end{align}
where $f: \mathcal{X} \times \Theta_t \to \mathbb{R}^{K}$ is a differentiable scoring function whose dependence on $\boldsymbol{\theta}_t$ is determined by which parameters are adapted at test time. The simplest case is when $f$ is linear in $\boldsymbol{\theta}_t$. That is the case, for example, of Bayesian logistic regression or methods that adapt only a linear classification head on top of a frozen feature extractor \cite{schirmer_temporal_2025}. More commonly in modern TTA, $f$ is nonlinear in $\boldsymbol{\theta}_t$. Examples of this include adapting internal parameters of a neural feature map, such as normalization statistics \cite{wang2021tent, zhang2022memo}, and injecting auxiliary parameters into a frozen feature map via low-rank coupling \cite{kojima2025lora}. Figure \ref{fig:structural_cases} summarizes the predictive model possibilities together with the adaptation parameter initialization.

\subsection{\texorpdfstring{Observation model $p(\mathcal{X}_t | \boldsymbol{\theta}_t)$}{Observation model}}

As mentioned earlier, due to the absence of labels, most of the works on TTA focus on optimizing the model's parameters based on a loss function of choice, generally relying on \emph{entropy minimization} and its variants \cite{grandvalet2004semi, marsden2024universal, zhou2021bayesian}. Modelling the observation model using a loss-based component as a pseudo-likelihood is well justified in the \emph{Gibbs posterior} literature \cite{bissiri2016general, knoblauch2022optimization}. We follow this approach and define the observation model as
\begin{align}
    p(\mathcal{X}_t | \boldsymbol{\theta}_t) = \exp \left( -\beta_t \ell(\mathcal{X}_t, \boldsymbol{\theta}_t) \right)
\end{align}
where $\beta_t > 0$ is a temperature parameter that controls the trade-off between trusting the prior and fitting the new data. Note that this expression is, in general, not a joint generative model due to the absence of the label sequence $\{ \mathcal{Y}_t \}_{t \in \mathbb{N}}$. It can be thought of as a scoring function that assigns a posterior weight to any parameter configuration given the observed test sequence. The standard supervised likelihood of the strictly Bayesian approach is replaced with an exponentiated loss while preserving the consistency of the update. This leaves the loss as a modelling choice that determines, in a decision theoretic approach, how each unlabeled test point $\mathbf{x}_t \in \mathcal{X}_t$ provides information about $\boldsymbol{\theta}_t$ according to a specific measure of fit, as justified in the generalized Bayesian framework \cite{bissiri2016general, knoblauch2022optimization}. 

The choice of the loss function for the Gibbs posterior is a central modelling decision, as it determines what information each unlabeled test point $\mathbf{x}_t$ provides about $\boldsymbol{\theta}_t$. Different choices correspond to different structural assumptions in the feature space and different probabilistic interpretations. Table \ref{tab:loss_functions_table} shows several types of losses that serve as a basis for most TTA works in the field. Note that for all of them, $\ell(\mathcal{X}_t;\boldsymbol{\theta}_t) = \frac{1}{N_t} \sum_{\mathbf{x}_t \in \mathcal{X}_t} \ell(\mathbf{x}_t;\boldsymbol{\theta}_t)$

\begin{table}[h!]
    \centering
    \footnotesize
    \renewcommand{\arraystretch}{1.5}
    \begin{tabular}{c | c }
         $\ell(\mathbf{x}_t;\boldsymbol{\theta}_t)$ & Description   \\ \hline 
         $- \sum_k p(\mathbf{y}_t = k | \mathbf{x}_t, \boldsymbol{\theta}_t) \cdot \log p(\mathbf{y}_t = k | \mathbf{x}_t, \boldsymbol{\theta}_t)$ & Entropy minimization \cite{grandvalet2004semi, wang2021tent, zhou2021bayesian} \\
         $- \sum_k p(\mathbf{y}_t = k | \mathbf{x}_t, \boldsymbol{\theta}_t) \cdot \log\left( \frac{p(\mathbf{y}_t = k | \mathbf{x}_t, \boldsymbol{\theta}_t)}{1 - p(\mathbf{y}_t = k | \mathbf{x}_t, \boldsymbol{\theta}_t)} \right)$  & Soft-likelihood ratio \cite{marsden2024universal, lee2024continual} \\
         $-\log p(\mathbf{\hat{y}}_t | \mathbf{x}_t, \boldsymbol{\theta}_t)$, $\mathbf{\hat{y}}_t = \arg \max_k p(\mathbf{y}_t = k | \mathbf{x}_t, \boldsymbol{\theta}_t)$ & Pseudo-label \cite{goyal2022test} \\
         $\ell_{SS}(\mathbf{x}_t; \boldsymbol{\theta}_t)$ & Self-supervised \cite{sun2020test, liu2021ttt++} 
    \end{tabular}
    \caption{Basic losses in TTA}
    \label{tab:loss_functions_table}
\end{table}
In the case in which the loss is the \emph{negative log marginal likelihood},  and $\beta_t = 1$, we recover the proper Bayesian posterior with target $\mathbf{y}_t$. This, of course, would require the specification of a class-conditional generative model characterized by $\boldsymbol{\theta}_t$, as in the prototype-based case \cite{iwasawa2021test, schirmer_temporal_2025}. In other cases, the loss will be a surrogate that does not correspond with a proper likelihood, with $\beta$ controlling for the strength of the data term relative to the prior. \emph{Entropy minimization} and \emph{soft-likelihood ratio} are based on the low-density separation assumption, the latter being more robust as it maintains non-vanishing gradients for confident predictions, preventing the frequent convergence collapse in entropy minimization. The \emph{pseudo-label} loss uses predictions as true labels, while the \emph{self-supervised loss} uses data augmentation techniques to enforce robustness, e.g, image transformations or adversarial perturbations \cite{}. Note that losses can be classified based on whether they are \emph{self-referential} or not, that is, whether they use the model's current output $p(\mathbf{y}_t | \mathbf{x}_t, \boldsymbol{\theta}_t)$ or not. These kinds of losses may lead to instabilities or collapse due to the moving-target problem, in which as $\boldsymbol{\theta}_t$ evolves, so does the loss landscape.

\paragraph{Choice of $\beta_t$}
As mentioned earlier, $\beta_t$ calibrates how much data information is meant to update the prior. A straightforward choice for this parameter is to set it to a constant scalar, that is, $\beta_t = \beta$, which can be selected as a hyperparameter of the model. However, this assumes that all samples carry equal information about $\boldsymbol{\theta}_t$, which is rarely the case in TTA. 

A more adaptive choice is to modulate $\beta_{t,i}$, a weight parameter per sample $i$ in the batch at time $t$, based on the model's current predictive confidence. Under \emph{reliability weighting}, we define the complement of the normalised predictive entropy $w_{t,i} = 1 - H \left( p(\mathbf{y}_t^i | \mathbf{x}_t^i, \boldsymbol{\theta}_t) \right) /\log C \in [0, 1]$, with $C$ the number of classes, and set $\beta_{t,i} = \beta \cdot w_{t,i}$. The motivation is that samples on which the model is already uncertain yield noisier pseudo-targets, and should arguably contribute less precision to the update. With this scheme, $\beta$ acts as a ceiling on the per-sample observation precision, and the total information injected in a step decreases naturally when the batch is collectively uncertain.

\subsection{\texorpdfstring{Transition model $p(\boldsymbol{\theta}_t | \boldsymbol{\theta}_{t-1})$}{Transition model}}

Once we have characterized the measurement function and the learning procedure, we will carry on to characterize the parameter transition model, which is the last element needed for computing the parameter prior and posterior at each time step $t$.

The transition function may be written as a Gaussian over the parameters of the previous step posterior, having the expression
\begin{align}
    \boldsymbol{\theta}_{t} = \mathbf{G}_t \boldsymbol{\theta}_{t-1} + \mathbf{c}_t + w_t \qquad w_t \sim \mathcal{N}(0, \mathbf{W}_t)
\end{align}
for some state matrix $\mathbf{G}_t$, vector $\mathbf{c}_t$, and covariance matrix $\mathbf{W}_t$. This construction of the transition function is standard in the dynamic linear model (DLM) literature \cite{west1997bayesian}, in which $\mathbf{G}_t$ encodes structural assumptions about the deterministic component of parameter dynamics. When the feature space has a direct physical interpretation, the addition of structural components through $\mathbf{G}_t$ is very meaningful. For example, a local linear trend model can encode acceleration drift, or a seasonal component can encode periodicity or oscillatory dynamics. If $\boldsymbol{\theta}_t$ parametrizes quantities with a direct relationship to the data-generating process, then these interpretations are fully justified. 

\begin{example}
Figure \ref{fig:effects_components} shows the effect on 1D linear regression that different structural components of $\mathbf{G}$ have on the parameter prior. We plot the parameter dynamics only — no data are observed — and fix an initial
intercept $\boldsymbol{\theta}_0 = 0$ and an initial slope $\boldsymbol{\theta}_1 = \tfrac{1}{2}$. The random walk causes the intercept to drift unboundedly due to accumulated noise. The polynomial trend does likewise, but with momentum from the hidden velocity state. The seasonal component oscillates periodically around the initial value. The Ornstein Uhlenbeck process confines both parameters to a stationary distribution centered at their initial values, with the degree of confinement controlled by $\rho$.

\begin{figure}[h!]
    \centering
    \includegraphics[width=\linewidth]{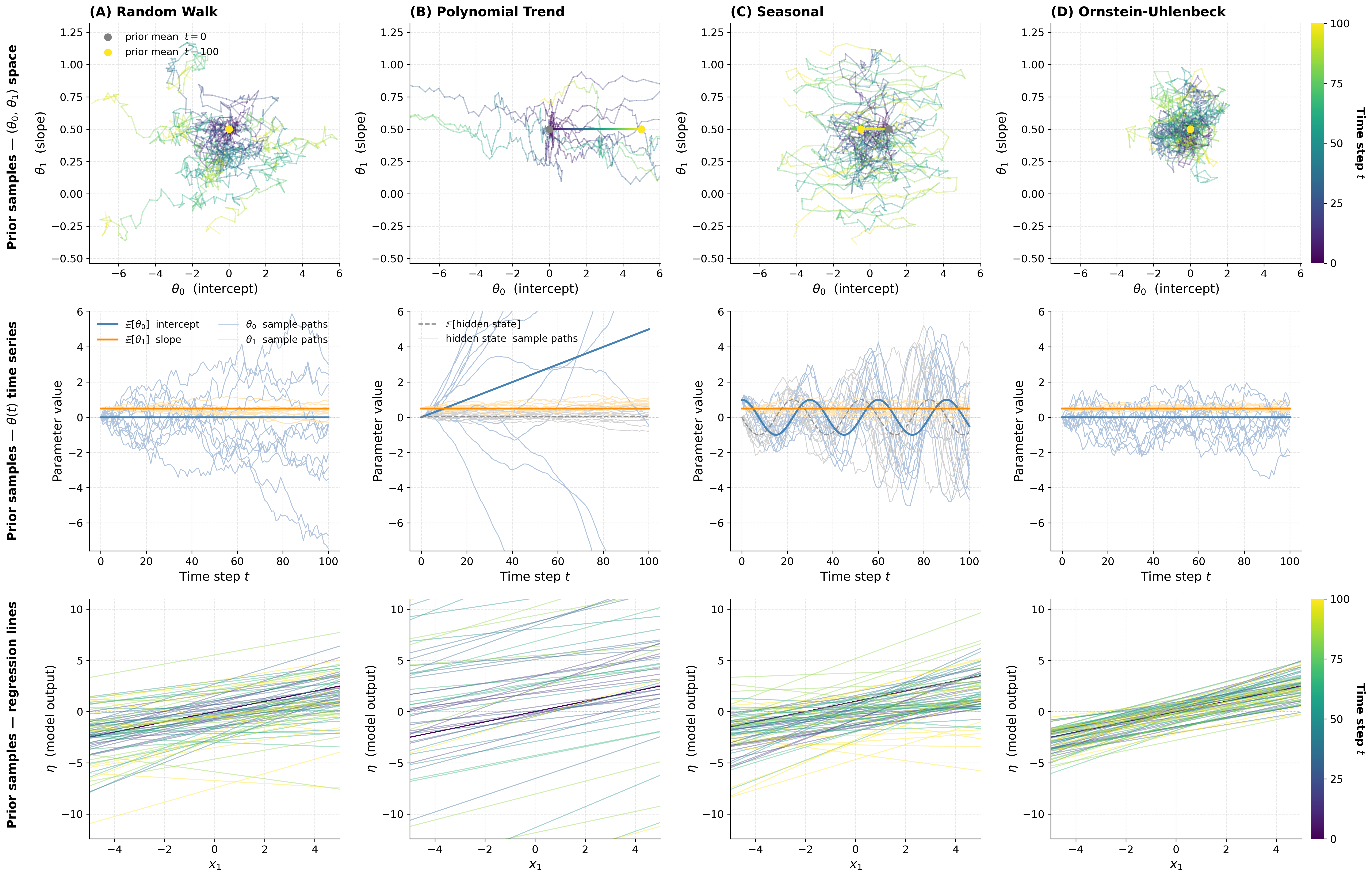}
    \caption{Prior parameter dynamics induced by four choices of transition matrix
             $\mathbf{G}$.
             \emph{Top row}: sample paths and prior mean in $(\boldsymbol{\theta}_0, \boldsymbol{\theta}_1)$
             parameter space, coloured by time.
             \emph{Middle row}: marginal time series of each parameter.
             \emph{Bottom row}: induced prior over regression lines
             $\eta = \boldsymbol{\theta}_1 x_1 + \boldsymbol{\theta}_0$.}
    \label{fig:effects_components}
    \end{figure}
\end{example}

When $\boldsymbol{\theta}_t$ are general discriminative parameters in a complex representation space without direct physical intuition, a structured $\mathbf{G}_t$ would be difficult to justify. The random walk $\mathbf{G}_t=I$ does not assume the direction of drift and is the default when the nature of the distribution shift is unknown. Source mean reversion, where $\mathbf{G}_t= \rho I$ and $\mathbf{c}_t = (1 - \rho)\boldsymbol{\mu}_0$ with $\rho \in [0,1]$, encodes the belief that adaptation should decay without sustained evidence towards the source mean $\boldsymbol{\mu}_0$, preventing unbounded drift from the source parameter distribution. 

The matrix $\mathbf{W}_t$ controls the speed and geometry of the stochastic parameter drift and is the more consequential design choice, with three main structural options. The first one is \emph{discounting} \cite{west1997bayesian}, in which $\mathbf{W}_t  = \frac{\delta - 1}{\delta} \boldsymbol{\Sigma}_{t-1}$, with $\delta \in (0,1]$ is a hyperparameter and where $\boldsymbol{\Sigma}_{t-1}$ is the covariance matrix of the previous-step posterior. Notice that this choice reduces computations of a new matrix while inflating the accumulated posterior by $1/{\delta}$ before seeing the data, becoming less confident before the next temporal step. Another option is the full specification of a matrix $\mathbf{W}_t$, the simplest example being isotropic noise $\mathbf{W}_t = \sigma^2 I$, which corresponds to uniform adaptation across all parameter directions, and is equivalent to the implicit process noise of a fixed learning rate in point-wise optimization methods. Finally, we could also consider dropping the covariance $\boldsymbol{\Sigma}_{t-1}$ completely, that is $\mathbf{W}_t = \widehat{\mathbf{W}}_t - \boldsymbol{\Sigma}_{t-1} $, and letting $\widehat{\mathbf{W}}_t = \sigma^2 I$ be the only curvature component. This option discards any previous curvature information and may deal better with large covariate shifts. The scale $\sigma^2$ is not directly identifiable from unlabeled test data alone and, thus, is left as a hyperparameter.

\section{Inference}
After building and exemplifying the necessary blocks, we focus on inference methods that estimate the posterior density of the parameters. We introduce two methods that assume Gaussian posteriors: linear-Gaussian approximations via linearization, and variational inference.

For full inference, we require computing the current-step prior distribution, which involves the previous-step posterior $p(\boldsymbol{\theta}_{t-1} | \mathcal{X}_{1:t-1}, \mathcal{D}_S)$ and the transition function $p(\boldsymbol{\theta}_t | \boldsymbol{\theta}_{t-1})$,
\begin{align}\label{eq:current-step-prior}
    p(\boldsymbol{\theta}_t| \mathcal{X}_{1:t-1}, \mathcal{D}_S) &=  \int p(\boldsymbol{\theta}_t| \boldsymbol{\theta}_{t-1}) \times p(\boldsymbol{\theta}_{t-1}| \mathcal{X}_{1:t-1}, \mathcal{D}_S)  \ d\boldsymbol{\theta}_{t-1} 
\end{align}
To simplify inference and retain conjugacy, we may
approximate the previous-step posterior by a Gaussian $p(\boldsymbol{\theta}_{t-1} | \mathcal{X}_{1:t-1}, \mathcal{D}_S) \approx \mathcal{N}(\boldsymbol{\mu}_{t-1}, \boldsymbol{\Sigma}_{t-1})$. Then, as the transition function has been defined also as a Gaussian $p(\boldsymbol{\theta}_{t}| \boldsymbol{\theta}_{t-1})  = \mathcal{N}(\mathbf{G}_t \boldsymbol{\theta}_{t-1} + \mathbf{c}_t,  \mathbf{W}_t)$, the current-step prior will have the expression
\begin{align}\label{eq:current-step-prior-gaussian}
    p(\boldsymbol{\theta}_t | \mathcal{X}_{1:t-1}, \mathcal{D}_S) = \mathcal{N}(\boldsymbol{\mu}_{t|t-1}, \boldsymbol{\Sigma}_{t|t-1}),
\end{align}
where $\boldsymbol{\mu}_{t|t-1} = \mathbf{G}_t \boldsymbol{\mu}_{t-1} + \boldsymbol{c}_t$ and $\boldsymbol{\Sigma}_{t|t-1} = \boldsymbol{\Sigma}_{t-1} + \mathbf{W}_t$. Thus, the current-step prior will also be Gaussian, and this property will be leveraged by the following inference methods. 

\subsection{Linear-Gaussian approximation via Taylor expansion of the loss}
Given that \ref{eq:current-step-prior-gaussian} is Gaussian, the Gibbs posterior \cite{bissiri2016general} at any time-step will have the expression
\begin{align}
    p(\boldsymbol{\theta}_t | \mathcal{X}_{1:t}, \mathcal{D}_S) &\propto p(\boldsymbol{\theta}_t | \mathcal{X}_{1:t-1}, \mathcal{D}_S) \exp (- \beta_t \ell_t(\mathcal{X}_t, \boldsymbol{\theta}_t))  \\
    &\propto \exp\left( -\frac{1}{2} (\boldsymbol{\theta}_t - \boldsymbol{\mu}_{t|t-1})^T \boldsymbol{\Sigma}_{t-1}^T(\boldsymbol{\theta}_t - \boldsymbol{\mu}_{t|t-1}) - \beta_t \ell_t(\mathcal{X}_t, \boldsymbol{\theta}_t)\right).
\end{align}
To get closed-form posterior, we can complete the square in the exponent by approximating the loss functions via a second-order Taylor expansion around the prior mean, that is,
\begin{align}
    \ell_t(\mathcal{X}_t, \boldsymbol{\theta}_t) \approx \ell_t(\mathcal{X}_t, \boldsymbol{\mu}_{t|t-1}) + g_t^T(\boldsymbol{\theta}_t - \boldsymbol{\mu}_{t|t-1}) + \frac{1}{2}(\boldsymbol{\theta}_t - \boldsymbol{\mu}_{t|t-1})^T  \mathbf{H}_t (\boldsymbol{\theta}_t - \boldsymbol{\mu}_{t|t-1})
\end{align}
where $g_t = \nabla \ell_t(\mathcal{X}_t, \boldsymbol{\theta}_t)|_{\boldsymbol{\theta}_t=\boldsymbol{\mu}_{t|t-1}}$ and $H_t = \nabla^2 \ell^T_t(\mathcal{X}_t, \boldsymbol{\theta}_t)|_{\boldsymbol{\theta}_t=\boldsymbol{\mu}_{t|t-1}}$. This results in a Gaussian with parameters
\begin{align}\label{eq:update_linloss}
    \begin{cases}
        \boldsymbol{\mu}_t &= \boldsymbol{\mu}_{t|t-1} - \beta_t\boldsymbol{\Sigma}_t g_t \\
        \boldsymbol{\Sigma}_t^{-1} &= \boldsymbol{\Sigma}_{t|t-1}^{-1} + \beta_t \mathbf{H}_t
    \end{cases}
\end{align}
Notice that this approach is equivalent to adopting a Newton-Raphson optimization step. The precision matrix accumulates the local curvature of the loss, and the mean moves based on the local gradient pre-conditioned by that curvature.

One limitation of this approximation is that it evaluates the gradient and the Hessian only at a single point, the mean $\boldsymbol{\mu}_{t|t-1}$, which for highly non-convex parameter spaces may lead to overconfident updates.

\subsection{Variational Bayes}

The above issue leads to an alternative method for approximating the parameter posterior distribution through Variational Bayes (VB) \cite{bishop2006pattern}, which transforms an estimation problem into an optimization problem. The idea is to approximate the posterior using a candidate parametric distribution (usually a Gaussian), chosen by minimizing the Kullback-Leibler divergence between the candidate distribution and the true posterior. The problem involves finding $q(\boldsymbol{\theta}_t^*) \sim (\boldsymbol{\mu}^*_t, \boldsymbol{\Sigma}^*_t)$ such that
\begin{align}
    (\boldsymbol{\mu}^*_t, \boldsymbol{\Sigma}^*_t) = \arg \min_{\boldsymbol{\mu}_t, \boldsymbol{\Sigma}_t} D_{KL} \left( \mathcal{N}(\boldsymbol{\mu}_t, \boldsymbol{\Sigma}_t) || \exp (- \beta_t \ell_t(\mathcal{X}_t, \boldsymbol{\theta}_t)) p(\boldsymbol{\theta}_t | \mathcal{X}_{1:t-1}, \mathcal{D}_S) \right)
\end{align}
Note that the normalization constant of the posterior may be ignored, a huge computational benefit of VB. Further, VB does not require linearization by construction. 

However, VB also has some drawbacks, as it requires an iterative optimization loop to converge to a solution and does not benefit from the sequentiality of the problem, necessitating the execution of a different optimization loop at each time step.

\subsection{Bayesian online natural gradient}

We leverage the work by \citet{jones2024bayesian}, which solves the VB computation issue by using one-step Natural Gradient Descent on the variational parameters for sequential variational Bayes, proving that it yields exact Bayesian inference when models are conjugate and serves as a relaxation of the Bayesian update for the non-conjugate variational case. The method, which assumes the variational family is Gaussian, leads to the following update:
\begin{align} \label{eq:update_BONG}
    \begin{cases}
        \boldsymbol{\mu}_t &= \boldsymbol{\mu}_{t|t-1} - \beta_t\boldsymbol{\Sigma}_t  \mathbb{E}_{q(\boldsymbol{\theta}_t)}[ \nabla_{\boldsymbol{\theta}_t} \ell_t(\mathcal{X}_t, \boldsymbol{\theta}_t)] \\
        \boldsymbol{\Sigma}_t^{-1} &= \boldsymbol{\Sigma}_{t|t-1}^{-1} + \beta_t  \mathbb{E}_{q(\boldsymbol{\theta}_t)}[\nabla^2_{\boldsymbol{\theta}_t} \ell_t(\mathcal{X}_t, \boldsymbol{\theta}_t)]
    \end{cases}
\end{align}
Notice how the update equations (\ref{eq:update_linloss}) and (\ref{eq:update_BONG}) look nearly identical. The difference is that Equation (\ref{eq:update_BONG}) evaluates an expected gradient and Hessian loss instead of point-estimates, thus leading to flatter and more robust minima compared to the linearization method. The integrals defined through these expectations are generally intractable, so they shall require an approximation. One option is using a Monte Carlo (MC) approximation, taking several independent samples from $q(\boldsymbol{\theta}_t)$ and estimating the expectation as the empirical mean of the inside terms evaluated at the selected samples. Another option is linearizing the loss over the previous step mean. This will exactly lead to Equation (\ref{eq:update_linloss}), thus showing the connection between the two approaches.

As said, we will use a Gaussian variational family to approximate the posterior distribution. However, different assumptions can be made regarding the covariance matrix $\boldsymbol{\Sigma}_t$. We will show examples of mean-field, diagonal-plus-low-rank, and full VI in the experiments section.

\subsection{Approximating the Hessian}

Both inference approaches require calculating the Hessian $\mathbf{H}_t$, which, to avoid numerical issues, is generally approximated using the positive-semidefinite \emph{Fisher Information Matrix (FIM)} \cite{martens2020new}
\begin{align}
    \mathbf{H}_t & \approx \mathbb{E} [ \nabla_{\boldsymbol{\theta}_{t}} \log p(\mathcal{X}_t|  \boldsymbol{\theta}_{t}) \cdot \nabla_{\boldsymbol{\theta}_{t}} \log p(\mathcal{X}_t|  \boldsymbol{\theta}_{t})^T ] \\
    & = \lambda^2 \mathbb{E} [ \left( \nabla_{\boldsymbol{\theta}_{t}} \ell_t(\mathcal{X}_t, \boldsymbol{\theta}_t) - \mathbb{E} [ \nabla_{\boldsymbol{\theta}_{t}} \ell_t(\mathcal{X}_t, \boldsymbol{\theta}_t) \right)  \cdot \left( \nabla_{\boldsymbol{\theta}_{t}} \ell_t(\mathcal{X}_t, \boldsymbol{\theta}_t) - \mathbb{E} [ \nabla_{\boldsymbol{\theta}_{t}} \ell_t(\mathcal{X}_t, \boldsymbol{\theta}_t) \right) ^T ] \\
    & = \lambda^2 \text{Var}[ \nabla_{\boldsymbol{\theta}_{t}} \ell_t(\mathcal{X}_t, \boldsymbol{\theta}_t) ],
\end{align}
 where expectations are calculated with respect to $p(X_t | \boldsymbol{\theta}_t)$. Since we cannot sample from this distribution directly, we may approximate the expectation via the empirical distribution, resulting in the \emph{empirical FIM}
 \begin{align}
     \mathbf{\hat{H}}_t = \frac{\lambda^2}{N_t} \sum_{i=1}^{N_t} \left( \nabla_{\boldsymbol{\theta}_{t}} \ell_t(\mathcal{X}^i_t, \boldsymbol{\theta}_t) - \overline{\mathbf{g}}_t  \right)  \cdot \left( \nabla_{\boldsymbol{\theta}_{t}} \ell_t(\mathcal{X}^i_t, \boldsymbol{\theta}_t) - \overline{\mathbf{g}}_t \right)^T,
 \end{align}
 where $\overline{\mathbf{g}}_t = \frac{1}{N_t} \sum_{i=1}^{N_t} \nabla_{\boldsymbol{\theta}_{t}} \ell_t(\mathcal{X}^i_t, \boldsymbol{\theta}_t)$
 
 Even calculating the FIM becomes computationally involved when the number of parameters is large, as it is a matrix of size $d \times d$, where $d$ is the number of parameters. However, modern computational methods to approximate and reduce this burden exist, based on factor approximations of this matrix, such as diagonal factorization, Kronecker-factored approximate curvature methods and low-rank factorizations \cite{daxberger2021laplace}.

\section{Prediction}
At each test step $t$, predictions are obtained by marginalising the generative model over the adapted posterior
\begin{align}
p(Y | \mathcal{X}_{1:t}, \mathcal{D}_S) = \int p(Y | \mathcal{X}_t, \boldsymbol{\theta}_t) \cdot p(\boldsymbol{\theta}_t | \mathcal{X}_{1:t}, \mathcal{D}_S) \, d\boldsymbol{\theta}_t, 
\end{align}
where $p(\boldsymbol{\theta}_t | \mathcal{X}_{1:t}, \mathcal{D}_S) = \mathcal{N}(\boldsymbol{\mu}_t, \boldsymbol{\Sigma}_t)$ is the Gaussian posterior over adaptation parameters after incorporating all past unlabelled test data and the source prior. This integral is intractable in general and needs to be approximated. Three main approximation methods exist:
\begin{itemize}
    \item The \emph{MAP approximation} substitutes the posterior mean for the full distribution, $p(Y | \mathcal{X}_t, \mathcal{D}_S) \approx p(Y | \mathcal{X}_t, \boldsymbol{\mu}_t)$, discarding all posterior uncertainty.
    \item The \emph{Monte Carlo approximation} draws $S$ samples $\boldsymbol{\theta}_t^{(s)} \sim \mathcal{N}(\boldsymbol{\mu}_t, \boldsymbol{\Sigma}_t)$ and estimates the predictive as $p(Y | \mathcal{X}_t, \mathcal{D}_S) \approx \frac{1}{S} \sum_s p(Y \mid \mathcal{X}_t, \boldsymbol{\theta}_t^{(s)})$, an unbiased estimator that converges as $S \to \infty$.
    \item The \emph{linearised approximation} yields a deterministic, closed-form predictive that propagates the full posterior covariance at the cost of a single Jacobian-vector product per class. This is cheaper than Monte Carlo and, unlike the MAP estimate, is sensitive to posterior uncertainty. Let $f(\mathcal{X}_t; \boldsymbol{\theta}_t)$ denote the logit function of Equation (\ref{eq:predictive_model}). A first-order Taylor expansion of $f$ around the posterior mean $\boldsymbol{\mu}_t$ gives
    \begin{align}
    f(\mathcal{X}_t; \boldsymbol{\theta}_t) \approx f(\mathcal{X}_t; \boldsymbol{\mu}_t) + \mathbf{J}(\mathcal{X}_t) (\boldsymbol{\theta}_t - \boldsymbol{\mu}_t),
    \end{align}
    where $\mathbf{J}(\mathcal{X}_t) = \partial f / \partial \boldsymbol{\theta}_t \big|_{\boldsymbol{\theta}_t = \boldsymbol{\mu}_t}$ is the Jacobian of the logits with respect to the adaptation parameters. We write $\mathbf{J}_c(\mathcal{X}_t) \in \mathbb{R}^{|\boldsymbol{\theta}_t|}$ for the (column) gradient of the $c$-th logit. Since $\boldsymbol{\theta}_t \sim \mathcal{N}(\boldsymbol{\mu}_t, \boldsymbol{\Sigma}_t)$, the linearisation induces a Gaussian distribution over $f(\mathcal{X}_t; \boldsymbol{\theta}_t)$ with per-class predictive variance
    \begin{align}
        \sigma_c^2 = \mathbf{J}_c(\mathcal{X}_t)^\top \boldsymbol{\Sigma}_t \mathbf{J}_c(\mathcal{X}_t).
    \end{align}
    The expected softmax under a Gaussian over logits has no closed form, so we apply the probit approximation \cite{bishop2006pattern} per class and renormalise, giving
    \begin{align}
    p(Y | \mathcal{X}_{1:t}, \mathcal{D}_S) \approx \operatorname{Categorical}\left(\operatorname{Softmax}\left(\boldsymbol{\kappa} \odot f(\mathcal{X}_t; \boldsymbol{\mu}_t)\right)\right),
    \end{align}
    where $\kappa_c = \left(1 + \tfrac{\pi}{8} \sigma_c^2\right)^{-1/2} \in (0,1]$.
\end{itemize}

\section{Experiments}
\subsection{Linear model}
\begin{example}[2D logistic regression with covariate shift under TTA]
    Consider a continuous stream of 2D data belonging to two classes and subject to a diagonal covariate shift. The drift is continuous, but its exact direction is unknown a priori. Hence, we decouple our parameter dynamics using a Dynamic Generalized Linear Model (DGLM) architecture. 
    
    For this, we define our state vector as $\boldsymbol{\theta}_t = [\boldsymbol{\theta}_{t,poly}, \boldsymbol{\theta}_{t,reg}^T]^T \in \mathbb{R}^{3}$, where $\boldsymbol{\theta}_{t,poly} \in \mathbb{R}$ represents the model bias, and $\boldsymbol{\theta}_{t,reg} \in \mathbb{R}^2$ represents the spatial regression weights. To encourage spatial stability while allowing the boundary to shift dynamically, we model the bias as a random walk (a zero-order polynomial trend) and the spatial weights as an Ornstein-Uhlenbeck (OU) process. 

    Using the prior precision (inverse of covariance) of the spatial weights $\Lambda_{\boldsymbol{\theta}_{t,reg}} = \boldsymbol{\Sigma}_{0, \boldsymbol{\theta}_{t,reg}}^{-1}$ and a mean-reversion rate $\eta = 0.001$, we assemble a time-constant block-diagonal state transition matrix $G$ and affine drift vector $c$
    \begin{align}
        \mathbf{G}_t&= \begin{pmatrix}
            1 & \mathbf{0}^T \\
            \mathbf{0} & I_2 - \eta \Lambda_{\boldsymbol{\theta}_{0,reg}}
        \end{pmatrix}, &
        c &= \begin{pmatrix}
            0 \\
            \eta \Lambda_{\boldsymbol{\theta}_{0,reg}} \boldsymbol{\mu}_{\boldsymbol{\theta}_{0,reg}}
        \end{pmatrix}
    \end{align}
    Following a standard supervised initialization to establish our prior $\mathcal{N}(\boldsymbol{\mu}_0, \boldsymbol{\Sigma}_0)$, we enter into the unsupervised Test-Time Adaptation (TTA) regime. At each time step $t$, we first project our beliefs forward. The mean transitions according to the affine DGLM dynamics, whereas the covariance expands by the block-diagonal system noise matrix $W$:
    \begin{align}
            p(\boldsymbol{\theta}_t | \mathbf{x}_{0:t-1}) &= \mathcal{N}(\boldsymbol{\mu}_{t|t-1}, \boldsymbol{\Sigma}_{t|t-1}) & &
            \begin{cases}
                \boldsymbol{\mu}_{t|t-1} &= \mathbf{G}_t\boldsymbol{\mu}_{t-1} + c\\
                \boldsymbol{\Sigma}_{t|t-1} &= G\boldsymbol{\Sigma}_{t-1}G^T + \mathbf{W_t}  
            \end{cases} 
    \end{align}
    Upon receiving an unlabeled data point $\mathbf{x}_t \in \mathbb{R}^2$, we construct the design vector $\mathbf{F}_t = [1, \mathbf{x}_{t}^T]^T$. We rely on entropy minimization to force the decision boundary to track the data drift. The exact gradient of the entropy loss $g_t = \nabla_{\boldsymbol{\theta}_t} \ell(\mathbf{x}_t, \boldsymbol{\theta}_t) |_{\boldsymbol{\theta}_t = \boldsymbol{\mu}_{t|t-1}}$ evaluates to:
    \begin{align}
        g_t = -\eta_t \big( \pi_t (1 - \pi_t) \big) \mathbf{F}_t,
    \end{align}
    where $\eta_t = \mathbf{F}_t^T \boldsymbol{\mu}_{t|t-1}$ is the predicted logit and  $\pi_t = \sigma(\eta_t)$ its corresponding probability. 
    
    By exponentiating this loss and performing a second-order Taylor expansion around the prior mean, we compute the Hessian $ \mathbf{H}_t$ (or its outer-product approximation). We update our Gaussian approximation using the standard Extended Kalman Filter (EKF) equations in information form:
    \begin{align}
            p(\boldsymbol{\theta}_t | \mathbf{x}_{0:t}) &\approx \mathcal{N}(\boldsymbol{\mu}_t, \boldsymbol{\Sigma}_{t}) & &
            \begin{cases}
                \boldsymbol{\Sigma}_{t}^{-1} &= \boldsymbol{\Sigma}_{t|t-1}^{-1} +  \mathbf{H}_t \\
                \boldsymbol{\mu}_t &= \boldsymbol{\mu}_{t|t-1} - \boldsymbol{\Sigma}_t g_t
            \end{cases} 
    \end{align}
    \begin{figure}[h!]
        \centering
        \includegraphics[width=1\linewidth]{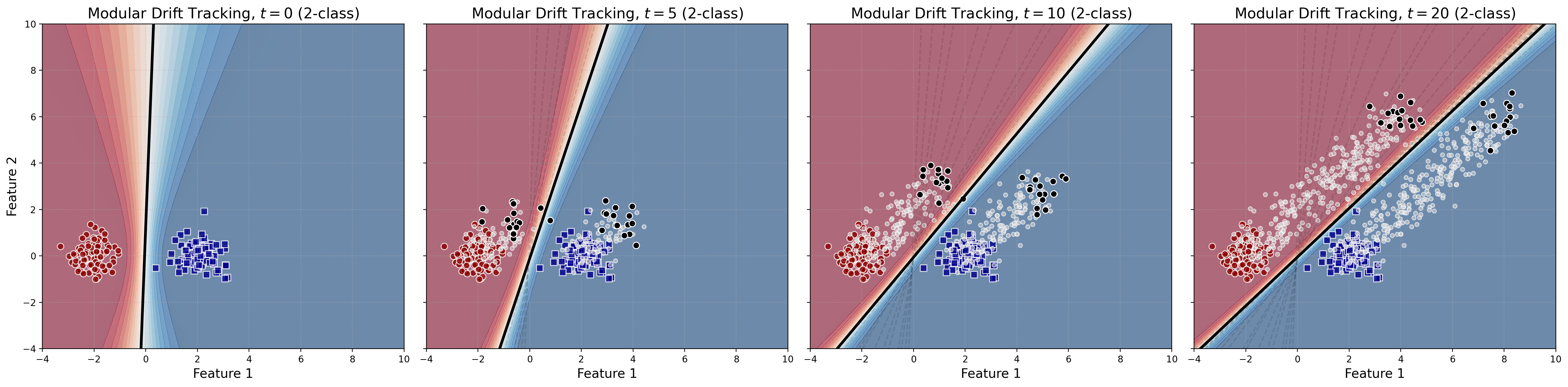}
        \caption{Classification boundary evolution}
        \label{fig:linear_model}
    \end{figure}
    We can observe how adaptation maintains predictive performance and detects the data separation region correctly. As time evolves, the predictor becomes more certain of the data distribution and uncertainty shrinks.
    
\end{example}

\subsection{Non-linear models}
\begin{example} [Two-moons shift dataset]
Consider a stream of 2D two-moon data, in which more data accumulates at opposite ends of each moon. Using a simple 4th-degree polynomial feature transformation and using logistic regression on the transformed features, we get via BONG inference the resulting predictive distributions shown in Figure \ref{fig:two:moons}. It is easy to see how the unlabeled data moves the decision boundary, maintaining a predictive posterior with good predictive performance. Uncertainty seems to shrink as time evolves, but it stays high in regions far from the observed data.
\begin{figure}[h!]
    \centering
    \includegraphics[width=1\linewidth]{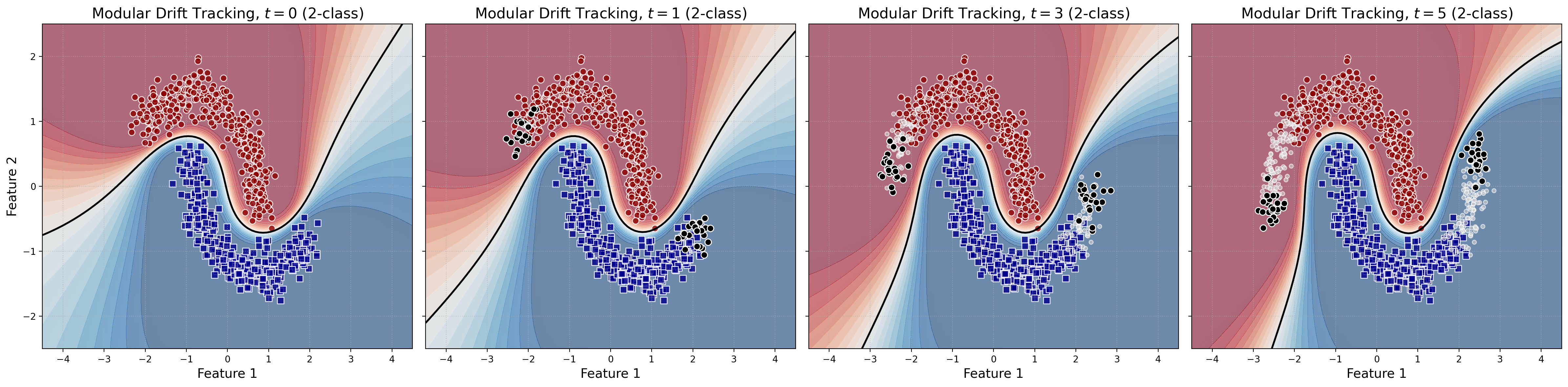}
    \caption{Classification boundary evolution}
    \label{fig:two:moons}
\end{figure}
\end{example}

\begin{example}[Three-class pinwheel shift dataset]
    Now consider a stream of 2D data with the shape of a pinwheel in which each of the blades belongs to a class. Using a 7th-degree polynomial feature transformation, and fitting the model using a Brier loss on Gaussian feature augmentations,  we get via BONG inference the resulting predictive distributions shown in Figure \ref{fig:two:moons}. We can see how the unlabeled data moves the decision boundaries, while all predictions also become more certain as time evolves.
\begin{figure}[h!]
    \centering
    \includegraphics[width=1\linewidth]{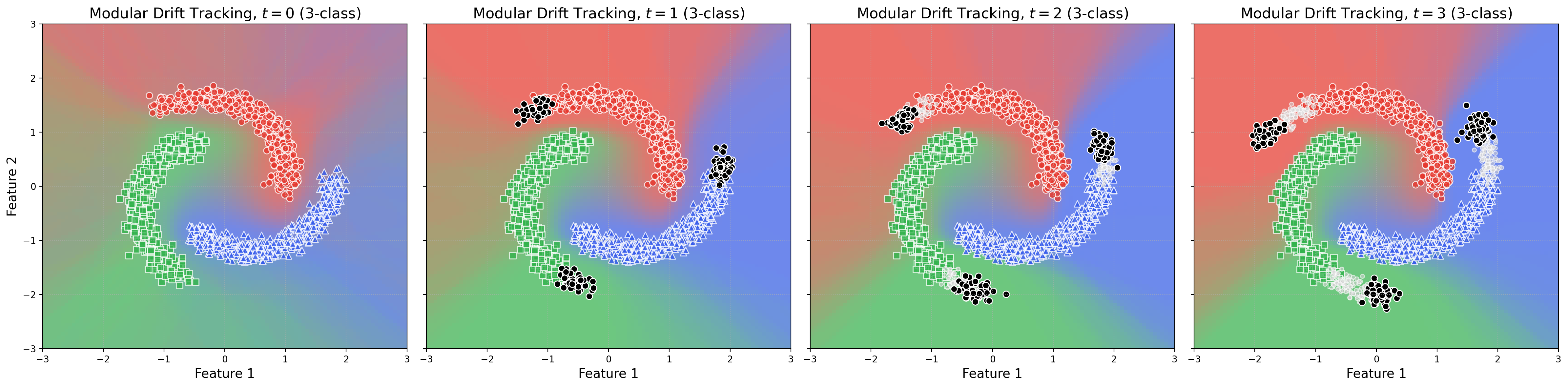}
    \caption{Classification boundary evolution}
    \label{fig:three-class-pinwheel}
\end{figure}
\end{example}

\subsection{Neural network TTA}
\subsubsection{Batch-norm adaptation}
\begin{example}[Online preconditioned gradient descent]
As a particular case of our framework, we can define an online preconditioned gradient descent algorithm that leverages the loss curvature information of the current step and discards the curvatures of previous steps. We define the parameters, following the previously used notation, as $\boldsymbol{\mu}_t = \boldsymbol{\mu}_{t|t-1} - \beta_t\boldsymbol{\Sigma}_t g_t$ and $\boldsymbol{\Sigma}_t^{-1} = W_r^{-1} + \beta_t H_t $. The covariance matrix does not use previous step information but uses noise-inflated curvature information of the current step. This approach may be seen as a curvature-informative point-based method for optimizing the parameters of the adapted model based on a degenerate distribution with point-mass at $\mu_t$.  

Table \ref{tab:cifar-c} shows the results of applying the online preconditioned gradient descent,  with information maximization as loss function and a constant $\beta_t = 2$, to the batch normalization parameters of a ResNet28-C pretrained on CIFAR10. The network is adapted independently to 15 different image corruptions as in the CIFAR10-C dataset, and Table \ref{tab:cifar-c} shows the average accuracy over the 15 corruptions. Additionally, it compares with other state-of-the-art methods solving the same problem. 

\begin{table}[h!]
\centering
  \begin{tabular}{|c|c|c|c|c|c|}
    \hline
    TTA Method & Source & BN test & TENT & EATA & Ours \\ \hline
    Avg Accuracy & 0.56 & 0.79 & 0.80 & 0.80 & $\mathbf{0.82}$ \\ \hline
  \end{tabular}
  \caption{CIFAR10-C: Average accuracy comparison with SOTA methods on corrupted CIFAR10 with 15 different corruptions, adapting batch normalization parameters of a ResNet-28. }
  \label{tab:cifar-c}
\end{table}
    
\end{example}

\section{Comparison with other TTA methods}

Most existing TTA methods can be recovered as special cases of the MAP limit of the proposed framework, with specific instantiations of the remaining components. TENT \citep{wang2021tent} corresponds to an entropy loss with constant $\beta$ and no source anchor; SHOT \citep{liang2020we} to an information-maximization objective; and pseudo-label methods \citep{goyal2022test} to cross-entropy against hard argmax targets. Several methods additionally incorporate source regularization: EATA's \citep{niu2022efficient} Fisher penalty and CoTTA's \citep{wang2022continual} stochastic parameter restore both serve to prevent unbounded drift from the source, playing an analogous role to the OU mean-reversion transition in our approach. In the MAP limit with a discount covariance propagation ($\mathbf{\Sigma}_{t|t-1} = \mathbf{\Sigma}_{t-1}/\delta$), the framework further reduces to an exponential moving average over past gradient information, connecting to EMA-based adaptation schemes. In all these mentioned cases, however, the posterior covariance is discarded after each update, and no uncertainty is propagated across steps. The general probabilistic framework we propose extends beyond this family by maintaining a full posterior over $\boldsymbol{\theta}_t$ across steps and encoding non-stationary parameter dynamics via the DGLM transition, neither of which is representable within gradient-based TTA methods.

\section{Conclusion}

This work presents a probabilistic model for test-time adaptation, whose characterization is modular and structured in clear building blocks. We separate the possible design choices based on data geometry assumptions from the distributional approximations needed to accelerate inference. We motivate the necessity for probabilistic approaches to test-time adaptation by showing that these can leverage model averaging and uncertainty estimation naturally, two elements that are crucial for better performance in test-time data-shifted scenarios. Increased predictive performance is shown in the linear and nonlinear simple scenarios, as well as for the adaptation of a pretrained neural network for corrupted image classification.

\pagebreak

\bibliography{sample}

\end{document}